\pdfoutput=1
\documentclass[11pt,a4paper]{article}
\usepackage[hyperref]{acl2020}
\usepackage{times}
\usepackage{latexsym}

\usepackage{inconsolata}
\usepackage{microtype}
\usepackage[fleqn]{amsmath}
\usepackage{amsfonts}
\usepackage{url}
\usepackage{booktabs}
\usepackage{multirow}
\usepackage{graphicx}
\usepackage{caption}
\usepackage{subcaption}
\usepackage{mathabx}
\usepackage{xparse}
\usepackage{xspace}
\usepackage{mathpartir}
\usepackage{tikz}
\usepackage{tikz-dependency}
\usepackage{etoolbox}
\usepackage{array,booktabs,ragged2e}
\usepackage{amsthm}
\usepackage{setspace}
\usepackage{bm}
\usepackage[T5,T1]{fontenc}
\usepackage[utf8]{inputenc}
\usepackage{xcolor}
\usepackage{colortbl}
\usepackage{tikz-dependency}
\usepackage{tikz}
\usepackage{lipsum}
\usepackage{algorithm}
\usepackage[noend]{algpseudocode}
\usepackage{tabularx}
\usepackage{CJKutf8}
\usepackage{refcount}
\usepackage{xstring}
\usepackage[title]{appendix}
\newcommand{\refsub}[1]{\StrRight{\getrefnumber{#1}}{1}}

\DeclareTextSymbolDefault{\OHORN}{T5}
\DeclareTextSymbolDefault{\UHORN}{T5}
\DeclareTextSymbolDefault{\ohorn}{T5}
\DeclareTextSymbolDefault{\uhorn}{T5}

\usepackage{gb4e}
\noautomath

\aclfinalcopy %

\def\aclpaperid{****} %

\title{
Extracting
Headless MWEs
from Dependency Parse Trees:\\
Parsing, Tagging, and Joint Modeling Approaches
}

\author{Tianze Shi\\
  Cornell University \\
  {\tt tianze@cs.cornell.edu} \\\And
  Lillian Lee\\
  Cornell University \\
  {\tt llee@cs.cornell.edu} \\
  }

\date{}

\mathchardef\mhyphen="2D

\newcommand{\reftab}[1]{Table~\ref{#1}}
\newcommand{\reffig}[1]{Fig.~\ref{#1}}
\newcommand{\refsec}[1]{\S\ref{#1}}

\newtheorem*{theorem*}{Theorem}

\newtheorem*{lemma*}{Lemma}

\newcolumntype{R}[1]{>{\RaggedLeft\arraybackslash}m{#1}}

\newcommand{\posscite}[1]{\citeauthor{#1}'s \citeyearpar{#1}}

\newcommand{\bivec}[1]{\ensuremath{\mathbf{#1}}}

\newcommand{\deprel}[1]{\textsf{#1}}

\newcommand{\treebank}[1]{\texttt{#1}}

\algrenewcommand\algorithmiccomment[1]{// {\itshape #1}}

\DeclareDocumentCommand{\trapezoid}{O{2.0} O{1.0} O{0.5} m m}{
    \begin{scope}[scale=0.9,thick]
        \draw[anchor=mid] (0, 0) -- (0, -{#2}) node[below=9pt,anchor=base] {\ensuremath{#4}} -- ({#1}, -{#2}) node [below=9pt,anchor=base] {\ensuremath{#5}} -- ({#1}, -{#3}) -- cycle;
    \end{scope}
}

\DeclareDocumentCommand{\righttriangle}{O{0.5} O{0.5} m m}{
    \begin{scope}[scale=0.9,thick]
        \draw[anchor=mid] (0, 0) -- (0, -{#2}) node[below=9pt,anchor=base] {\ensuremath{#3}} -- ({#1}, -{#2}) node [below right=9pt and 3pt,anchor=base] {\ensuremath{#4}} -- cycle;
    \end{scope}
}

\DeclareDocumentCommand{\lefttriangle}{O{0.5} O{0.5} m m}{
    \begin{scope}[scale=0.9,thick]
        \draw[anchor=mid] (0, -{#2}) node[below left=9pt and 3pt,anchor=base] {\ensuremath{#3}} -- ({#1}, -{#2}) node [below=9pt,anchor=base] {\ensuremath{#4}} -- ({#1}, 0) -- cycle;
    \end{scope}
}

\newcommand{\ch}[1]{\begin{CJK}{UTF8}{bkai}#1\end{CJK}}
\newcommand{\ja}[1]{\begin{CJK}{UTF8}{min}#1\end{CJK}}

\newcommand{\resnumber}[2]{\ensuremath{#1_{\pm #2}}}
\newcommand{\best}[2]{{\cellcolor{cyan!25}\ensuremath{\resnumber{\mathbf{#1}}{#2}}}}
\newcommand{\close}[2]{\ensuremath{\resnumber{\mathbf{#1}}{#2}}}

\begin{document}
\maketitle

\begin{abstract}
An interesting and frequent type of
multi-word expression (MWE) is
the headless MWE,
for which there 
are no true internal syntactic dominance relations;
examples include many named entities (``Wells Fargo'') and dates (``July 5, 2020'')
as well as certain productive constructions (``blow for blow'', ``day after day'').
Despite their special status and prevalence, current dependency-annotation
schemes require treating such flat structures as if they had internal syntactic heads,
and most current parsers handle them in the same fashion as
headed constructions.
Meanwhile, outside the context of
parsing,
taggers are typically used for identifying
MWEs,
but taggers might benefit from structural information.
We empirically compare these two common strategies---parsing and tagging---for
predicting flat
MWEs.
Additionally, we propose
an efficient joint decoding algorithm that combines scores from both strategies.
Experimental results on the MWE-Aware English Dependency Corpus and
on
six non-English dependency treebanks with frequent flat structures
show that:
(1) tagging is more accurate than parsing for identifying flat-structure MWEs,
(2) our joint decoder reconciles the two different views and,
for non-BERT features,
 leads to
 higher accuracies,
and (3) most of the gains result from feature sharing between the parsers and  taggers.

\end{abstract}

\section{Introduction}
\label{sec:intro}

\begin{figure}[t]
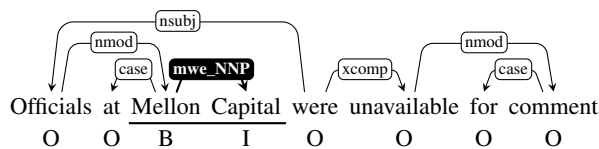

\small
\centering
    \hspace*{-10pt}
\begin{dependency}
\begin{deptext}
    Officials \& at \& Mellon \& Capital \& were \& unavailable \& for \& comment\\[3pt]
O \& O \& B \& I \& O \& O \& O \& O\\
\end{deptext}
    \depedge[edge height=6.5ex]{5}{1}{nsubj}
    \depedge[edge height=2ex]{3}{2}{case}
    \depedge[edge height=4.5ex]{1}{3}{nmod}
    \depedge[theme=night,edge height=2ex]{3}{4}{mwe\_NNP}
    \depedge[edge height=2ex]{5}{6}{xcomp}
    \depedge[edge height=2ex]{8}{7}{case}
    \depedge[edge height=4.5ex]{6}{8}{nmod}
    \node (ll) [below left of = \wordref{1}{3}, xshift=3pt, yshift=14pt] {};
    \node (rr) [below right of = \wordref{1}{4}, xshift=-2pt, yshift=14pt] {};
    \draw [-, thick, black] (ll) -- node[anchor=north] {} (rr);
\end{dependency}
\caption{
Dependency tree from the MWE-Aware English Dependency Corpus, imposing a ``head''
relationship between the words in the actually headless MWE {\sf Mellon Capital}.
Also shown are MWE BIO labels.
}
\label{fig:ptb-ex}

\end{figure}

Headless multi-word expressions (MWEs), including many named entities and certain productive constructions,
are frequent in natural language and are important to NLP applications.
In the context of dependency-based syntactic parsing, however, they pose an interesting representational challenge.
Dependency-graph formalisms for syntactic structure represent lexical items as
nodes and head-dominates-modifier/argument relations between lexical items
as directed arcs on the corresponding pair of nodes. Most words can be assigned
clear linguistically-motivated syntactic heads, but several frequently occurring phenomena do not
easily fit into this framework, including punctuation, coordinating
conjunctions, and
``flat'', or headless MWEs. While the proper treatment of
headless constructions in dependency formalisms remains debated
\citep{kahane-etal-2017-multi,gerdes+al:2018:mwesForUD},
many well-known
dependency treebanks
handle MWEs by giving their component
words a ``default head'', which is not indicative of a true dominance relation,
but rather as ``a tree encoding of a flat structure without a syntactic head'' \citep[pg. 213]{demarneffe-nivre19}.
\reffig{fig:ptb-ex} shows an example: the headless MWE {\sf Mellon Capital}
has its first word, {\sf Mellon},  marked as
the ``head'' of {\sf Capital}.

Despite the special status of flat structures in dependency tree annotations,
most state-of-the-art dependency parsers treat all annotated relations equally,
and thus do not distinguish between headed and headless constructions.
When
headless-span identification (e.g., as part of named-entity recognition (NER)) is the %
specific task at hand,
\underline{b}egin-chunk/\underline{i}nside-chunk/\underline{o}utside-chunk (BIO) tagging \citep{ramshaw-marcus-1995-text}
is
generally adopted.
It is therefore natural to ask whether \emph{parsers} are as 
accurate
as \emph{taggers}
in identifying these ``flat branches'' in dependency trees.
Additionally, since parsing and tagging represent two different views of the same underlying structures,
can \emph{joint decoding} that combines scores from the two modules
and/or \emph{joint training} under a multi-task learning (MTL) framework
derive more accurate models than parsing or tagging alone?

To facilitate answering these questions,
we introduce a joint decoder that finds the maximum sum of scores from both BIO tagging and parsing decisions.
The joint decoder incorporates a special deduction item representing continuous headless spans,
while retaining the cubic-time efficiency of projective dependency parsing.
The outputs are consistent structures across the tagging view and the parsing view.

We perform evaluation of the different strategies
on the MWE-Aware English Dependency Corpus
and
treebanks for
five additional languages from
the Universal Dependencies 2.2 corpus that have frequent multi-word headless constructions.
On average, we find taggers to be more accurate than parsers at this task,
providing $0.59\%$ ($1.42\%$) absolute higher F1 scores
with(out) pre-trained contextualized word representations.
Our joint decoder combining jointly-trained taggers and parsers further
improves the tagging strategy by $0.69\%$ ($1.64\%$) absolute.
This corroborates early evidence \cite{finkel-manning09} that joint modeling with parsing improves over NER.
We also show that neural representation sharing through MTL is an effective strategy,
as it accounts for a large portion of our observed improvements.
Our code is publicly available at \url{https://github.com/tzshi/flat-mwe-parsing}.
\begin{figure*}[!htp]
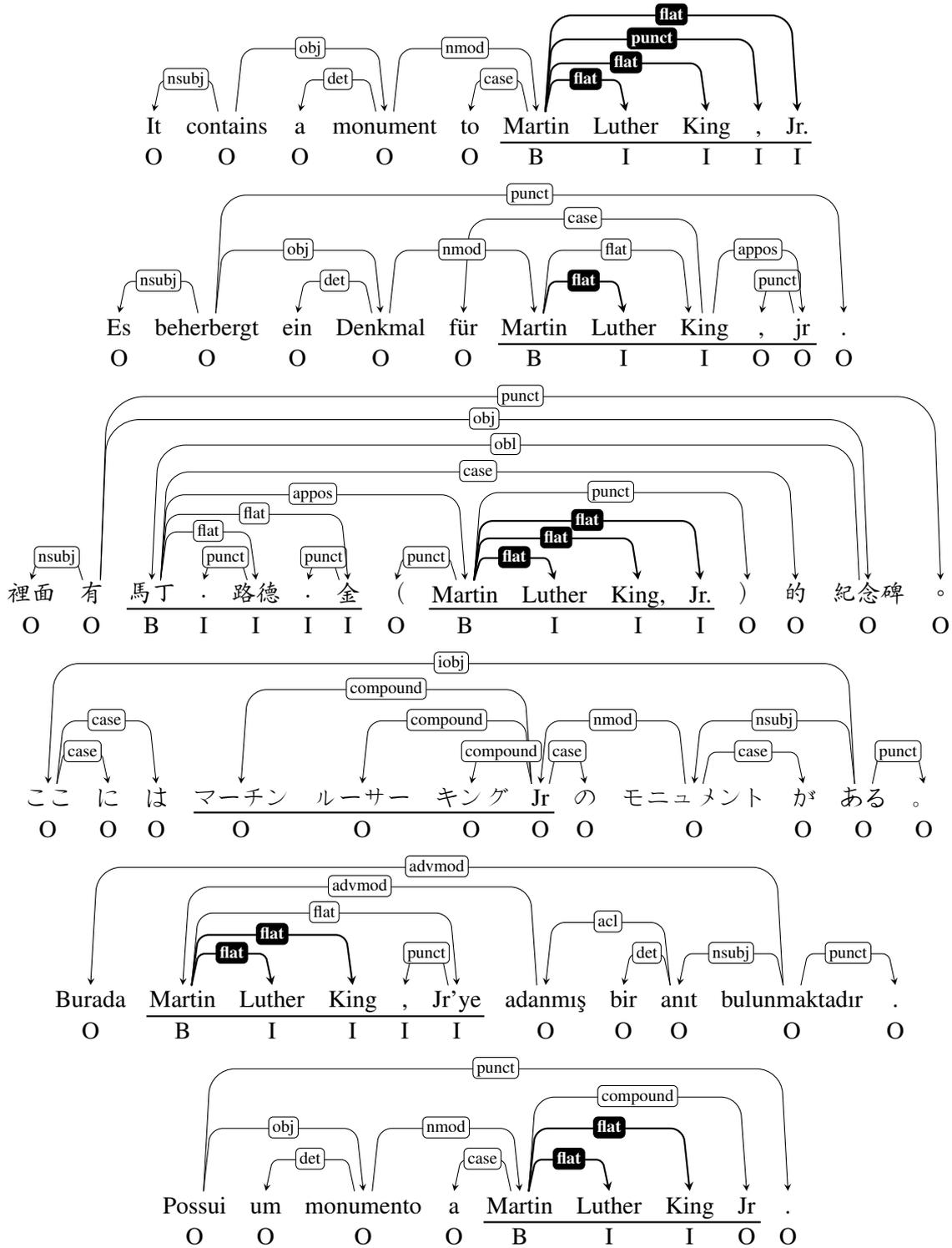

\centering

\begin{dependency}
\begin{deptext}[column sep=0.2cm]
    It \& contains \& a \& monument \& to \& Martin \& Luther \& King \& , \& Jr.\\[3pt]
O \& O \& O \& O \& O \& B \& I \& I \& I \& I\\
\end{deptext}
\depedge{2}{1}{nsubj}
\depedge{4}{3}{det}
\depedge{2}{4}{obj}
\depedge{6}{5}{case}
\depedge{4}{6}{nmod}
\depedge[theme=night]{6}{7}{flat}
\depedge[edge unit distance=2.2ex, theme=night]{6}{8}{flat}
\depedge[edge unit distance=2.2ex, theme=night]{6}{9}{punct}
\depedge[edge unit distance=2.2ex, theme=night]{6}{10}{flat}
\node (ll) [below left of = \wordref{1}{6}, yshift=13pt] {};
\node (rr) [below right of = \wordref{1}{10}, xshift=-10pt, yshift=13pt] {};
\draw [-, thick, black] (ll) -- node[anchor=north] {} (rr);
\end{dependency}

\begin{dependency}
\begin{deptext}[column sep=0.2cm]
    Es \& beherbergt \& ein \& Denkmal \& für \& Martin \& Luther \& King \& , \& jr \& .\\[3pt]
O \& O \& O \& O \& O \& B \& I \& I \& O \& O \& O\\
\end{deptext}
\depedge{2}{1}{nsubj}
\depedge{4}{3}{det}
\depedge{2}{4}{obj}
\depedge[edge start x offset=3pt]{8}{5}{case}
\depedge{4}{6}{nmod}
\depedge[theme=night]{6}{7}{flat}
\depedge[edge end x offset=-7pt]{6}{8}{flat}
\depedge{10}{9}{punct}
\depedge{8}{10}{appos}
\depedge[edge unit distance=1.2ex]{2}{11}{punct}
\node (ll) [below left of = \wordref{1}{6}, yshift=13pt] {};
\node (rr) [below right of = \wordref{1}{10}, xshift=-10pt, yshift=13pt] {};
\draw [-, thick, black] (ll) -- node[anchor=north] {} (rr);
\end{dependency}

\begin{dependency}
\begin{deptext}[column sep=0.2cm]
    \ch{裡面} \& \ch{有} \& \ch{馬丁} \& \ch{·} \& \ch{路德} \& \ch{·} \& \ch{金} \& \ch{（} \& Martin \& Luther \& King, \& Jr. \& \ch{）} \& \ch{的} \& \ch{紀念碑} \& \ch{。}\\[3pt]
O \& O \& B \& I \& I \& I \& I \& O \& B \& I \& I \& I \& O \& O \& O \& O\\
\end{deptext}
\depedge[edge unit distance=2.ex]{2}{1}{nsubj}
\depedge[edge unit distance=1.03ex]{15}{3}{obl}
\depedge[edge unit distance=2.ex]{5}{4}{punct}
\depedge[edge unit distance=2.2ex]{3}{5}{flat}
\depedge[edge unit distance=2.ex]{7}{6}{punct}
\depedge[edge unit distance=1.5ex]{3}{7}{flat}
\depedge[edge unit distance=2.ex]{9}{8}{punct}
\depedge[edge unit distance=1.3ex]{3}{9}{appos}
\depedge[edge unit distance=2.ex, theme=night]{9}{10}{flat}
\depedge[edge unit distance=1.9ex, theme=night]{9}{11}{flat}
\depedge[edge unit distance=1.8ex, theme=night]{9}{12}{flat}
\depedge[edge unit distance=2.ex]{9}{13}{punct}
\depedge[edge unit distance=0.9ex]{3}{14}{case}
\depedge[edge unit distance=1.13ex]{2}{15}{obj}
\depedge[edge unit distance=1.2ex]{2}{16}{punct}
\node (ll) [below left of = \wordref{1}{3}, xshift=5pt, yshift=14pt] {};
\node (rr) [below right of = \wordref{1}{7}, xshift=-10pt, yshift=14pt] {};
\draw [-, thick, black] (ll) -- node[anchor=north] {} (rr);
\node (ll2) [below left of = \wordref{1}{9}, xshift=0pt, yshift=14pt] {};
\node (rr2) [below right of = \wordref{1}{12}, xshift=-10pt, yshift=14pt] {};
\draw [-, thick, black] (ll2) -- node[anchor=north] {} (rr2);
\end{dependency}

\begin{dependency}
\begin{deptext}[column sep=0.2cm]
    \ja{ここ} \& \ja{に} \& \ja{は} \& \ja{マーチン} \& \ja{ルーサー} \& \ja{キング} \& \ja{Jr} \& \ja{の} \& \ja{モニュメント} \& \ja{が} \& \ja{ある} \& \ja{。} \\[3pt]
O \& O \& O \& O \& O \& O \& O \& O \& O \& O \& O \& O\\
\end{deptext}
\depedge[edge unit distance=1.1ex]{11}{1}{iobj}
\depedge{1}{2}{case}
\depedge{1}{3}{case}
\depedge{7}{4}{compound}
\depedge{7}{5}{compound}
\depedge{7}{6}{compound}
\depedge{9}{7}{nmod}
\depedge{7}{8}{case}
\depedge{11}{9}{nsubj}
\depedge{9}{10}{case}
\depedge{11}{12}{punct}
\node (ll) [below left of = \wordref{1}{4}, xshift=-5pt, yshift=14pt] {};
\node (rr) [below right of = \wordref{1}{7}, xshift=-10pt, yshift=14pt] {};
\draw [-, thick, black] (ll) -- node[anchor=north] {} (rr);
\end{dependency}

\begin{dependency}
\begin{deptext}[column sep=0.2cm]
    Burada \& Martin \& Luther \& King \& , \& Jr'ye \& adanmış \& bir \& anıt \& bulunmaktadır \& .\\[3pt]
O \& B \& I \& I \& I \& I \& O \& O \& O \& O \& O\\
\end{deptext}
\depedge[edge unit distance=1.2ex]{10}{1}{advmod}
\depedge[edge unit distance=1.8ex]{7}{2}{advmod}
\depedge[theme=night]{2}{3}{flat}
\depedge[edge unit distance=2.3ex, theme=night]{2}{4}{flat}
\depedge{6}{5}{punct}
\depedge[edge unit distance=1.65ex]{2}{6}{flat}
\depedge{9}{7}{acl}
\depedge{9}{8}{det}
\depedge{10}{9}{nsubj}
\depedge{10}{11}{punct}
\node (ll) [below left of = \wordref{1}{2}, xshift=0pt, yshift=13pt] {};
\node (rr) [below right of = \wordref{1}{6}, xshift=-4pt, yshift=13pt] {};
\draw [-, thick, black] (ll) -- node[anchor=north] {} (rr);
\end{dependency}

\begin{dependency}
\begin{deptext}[column sep=0.2cm]
    Possui \& um \& monumento \& a \& Martin \& Luther \& King \& Jr \& .\\[3pt]
O \& O \& O \& O \& B \& I \& I \& O \& O\\
\end{deptext}
\depedge{3}{2}{det}
\depedge{1}{3}{obj}
\depedge{5}{4}{case}
\depedge{3}{5}{nmod}
\depedge[theme=night]{5}{6}{flat}
\depedge[theme=night]{5}{7}{flat}
\depedge{5}{8}{compound}
\depedge[edge unit distance=1.4ex]{1}{9}{punct}
\node (ll) [below left of = \wordref{1}{5}, xshift=0pt, yshift=13pt] {};
\node (rr) [below right of = \wordref{1}{8}, xshift=-10pt, yshift=13pt] {};
\draw [-, thick, black] (ll) -- node[anchor=north] {} (rr);
\end{dependency}

\caption{
    An illustration of flat-structure annotation variation across treebanks:
    a set of parallel sentences, all containing the conceptually headless MWE ``Martin Luther King, Jr.'' (underlined),
    from UD 2.2 (treebank code \treebank{\_pud}) in English, German, Chinese, Japanese, Turkish, and Portuguese (top to bottom).
    The intent of this figure is not to critique particular annotation decisions, but to demonstrate the notation, concepts,
    and data extraction methods used in our paper.
    To wit: Highlights/black-background indicate well-formed flat-MWE tree fragments
    according to the principles listed in \refsec{sec:data}.
    BIO sequences are induced by the
    longest-spanning \deprel{flat} arcs.
    When there is a mismatch between the highlighted tree fragments
    and the BI spans---here, in the German, Chinese and Turkish examples---it is because
    the dependency trees do not fully conform to the UD annotation guidelines on headless structures.
}
\label{fig:ud-ex}

\end{figure*}

\section{Background on Headless Structures}
\label{sec:repr}

\begin{table*}[ht]
  \centering
    \begin{tabular}{m{12cm}|R{1.3cm}R{1cm}}
    \toprule
        \multirow{2}{*}{Treebank (Language)} & \multicolumn{2}{c}{\% of \deprel{flat}} \\
        & graphs $\downarrow$ & arcs \\
    \midrule
    \emph{19 treebanks with highest percentages}: \\
    \treebank{ko\_gsd} (Korean) & $67.84$ & $15.35$ \\
    \treebank{id\_gsd} (Indonesian) & $61.63$ & $9.39$ \\
    \treebank{ca\_ancora} (Catalan) & $41.11$ & $3.32$ \\
    \treebank{nl\_lassysmall} (Dutch) & $38.90$ & $5.87$ \\
    \treebank{ar\_nyuad} (Arabic) & $37.63$ & $2.19$ \\
    \treebank{es\_ancora} (Spanish),
    \treebank{sr\_set} (Serbian),
    \treebank{it\_postwita} (Italian),
    \treebank{pt\_bosque} (Portuguese),
    \treebank{pt\_gsd} (Portuguese),
    \treebank{fa\_seraji} (Persian),
    \treebank{de\_gsd} (German),
    \treebank{hu\_szeged} (Hungarian),
    \treebank{fr\_gsd} (French),
    \treebank{es\_gsd} (Spanish),
    \treebank{he\_htb} (Hebrew),
    \treebank{kk\_ktb} (Kazakh),
    \treebank{be\_hse} (Belarusian),
    \treebank{nl\_alpino} (Dutch)
    & $>20.00$ &  \\
    \midrule
    \ldots & \ldots \\
    \midrule
    \emph{12 treebanks without \deprel{flat} arcs}: \\
    \treebank{cs\_cltt} (Czech),
    \treebank{grc\_perseus} (Ancient Greek),
    \treebank{hi\_hdtb} (Hindi),
    \treebank{ja\_gsd} (Japanese),
    \treebank{ja\_bccwj} (Japanese),
    \treebank{la\_ittb} (Latin),
    \treebank{la\_perseus} (Latin),
    \treebank{no\_nynorsklia} (Norwegian),
    \treebank{swl\_sslc} (Swedish Sign Language),
    \treebank{ta\_ttb} (Tamil),
    \treebank{ur\_udtb} (Urdu),
    \treebank{vi\_vtb} (Vietnamese)
    & $0.00$ & $0.00$ \\
    \bottomrule
    \end{tabular}%
  \caption{
    The UD 2.2 training treebanks with highest and lowest percentage of \deprel{flat} arcs, out of 
    $90$ treebanks.
  }
  \label{tab:ud-basic-stats}%
\end{table*}%

A (multi-word) headless construction, or flat structure, is a span of lexical items that together
reference a single concept and where no component is a syntactically more plausible
candidate for
the span's head than any other component.
Examples are boldfaced in the following
English sentences.

\begin{exe}
\ex
    \emph{Within the scope of this paper:}
    \begin{xlist}
    \ex \label{eg:1b}ACL starts on {\bf July 5, 2020}.
    \ex \label{eg:1c}My bank is {\bf Wells Fargo}.
    \ex \label{eg:1a}The candidates matched each other {\bf insult for insult}. \citep{jackendoff:2008:constructionforconstruction}
    \end{xlist}
    \label{eg:1}
\end{exe}

(\ref{eg:1})\refsub{eg:1b}\
and
(\ref{eg:1})\refsub{eg:1c}\
show that dates and many named entities can be headless constructions,
suggesting that they are frequent.
Indeed,
in the MWE-Aware English Dependency Corpus \citep{kato+17},
nearly half of the sentences contain headless constructions,
75\% of which are named entities.
For comparison,
(\ref{eg:2}) shows examples of non-flat MWEs,
which are also interesting and important, but they are beyond the focus of our paper.
\begin{exe}
\ex \emph{Outside the scope of this paper:}
    \begin{xlist}
    \ex \label{eg:2a}{\bf congressman at large} \citep{sag02}  [head = ``congressman'']
    \ex \label{eg:2b} I have {\bf moved on}. [verb-particle construction, head = ``moved'']
    \ex \label{eg:2c} I {\bf take} your argument {\bf into account}. \citep{constant+17}  [light-verb construction, head = ``take'']
    \end{xlist}
    \label{eg:2}
\end{exe}

Returning to
headless MWEs, the choice of representation for headless spans depends on the task.
In \emph{named-entity recognition}, such spans are
often treated
as BIO tag sequences:\footnote{
In this paper, we adopt the original BIO tagset, which cannot properly represent discontinuous MWEs.
See \citet{schneider+14} for modified tagsets
providing such support.
}
for example, in \reffig{fig:ptb-ex}, ``Mellon'' is tagged as ``B'' and ``Capital'' is tagged as ``I''.
In \emph{dependency parsing}, where labeled dependency arcs are the only way to express a syntactic analysis
(short of treating MWEs as atomic lexical items, which would result in a chicken-and-egg problem)
is to impose arcs within the MWE's span.
Different corpora adopt different annotation conventions.
The MWE-Aware English Dependency Corpus uses the arc label \deprel{mwe\_NNP}, as shown in \reffig{fig:ptb-ex}.
The Universal Dependencies \cite[UD;][]{ud22}
annotation guidelines have all following tokens in such constructions
attached to the first one via
arcs labeled \deprel{flat},
a choice that is admittedly
``in principle arbitrary''.\footnote{\href{https://universaldependencies.org/u/dep/flat.html}{\nolinkurl{universaldependencies.org/u/dep/flat.html}}}
The frequency of flat structures across different treebanks varies according to
language, genre, and even  tokenization guidelines, among other factors.
\reftab{tab:ud-basic-stats} lists
the UD 2.2 treebanks
with the highest and lowest percentage of \deprel{flat} relations.
While the Korean treebank \treebank{ko\_gsd} (with the highest percentage) splits up most names into multiple tokens and connects them through \deprel{flat},
the Japanese treebank \treebank{ja\_gsd}
(no \deprel{flat}s at all)
treats all names as compound nouns, and
thus represents them as having internal structure without any indication
 that a special case  has occurred.\footnote{
Some flat structures can end up using other dependency labels such as \deprel{compound},
as a result of the fact that
many UD treebanks, including \treebank{ja\_gsd}, are automatically converted from non-UD style annotations.
The UD annotations depend on how detailed the original syntactic analyses are and the accuracies of the conversion algorithms.
}
\reffig{fig:ud-ex} shows examples from the UD parallel treebanks, illustrating the diversity of annotation for the same sentence rendered in different languages.

Overall, more than $20\%$ of the treebanks in the UD 2.2 collection have \deprel{flat} structures in more than $20\%$ of their training-set sentences.\footnote{
Measured on the $90$ treebanks with training splits.
}
Therefore,
a parsing approach taking into account the special status of headless structural representations
can potentially benefit models for a large number of languages and treebanks.

\subsection{Notation and Definitions}
Formally, given an $n$-word sentence $w=w_1, w_2, \ldots, w_n$,
we define its dependency structure to be
a graph
$G = (V, E)$.
Each node in $V$ corresponds to a word in the sentence.
Each (labeled) edge $(h, m, r) \in E$
denotes a syntactic relation labeled $r$ between the head word $w_h$ and modifier word $w_m$,
where $h,m\in\{0,1,\ldots,n\}$
and $0$ denotes the dummy root of the sentence.
Since we work with dependency treebanks, we require that
the edges in $E$ form a tree.
To represent a multi-word headless span $w_i, \ldots, w_j$,
all subsequent words in the span are attached to the beginning word $w_i$,
i.e., $\forall k \in \{i+1,\ldots, j\},
 (i, k, f) \in E$,
where $f$ is the special syntactic relation label
denoting headless structures
(\deprel{flat} in UD annotation).
Alternatively, one can also use a BIO tag sequence $T=(t_1, t_2, \ldots, t_n) \in \{\text{B},\text{I},\text{O}\}^n$ to
indicate the location of any headless spans within $w$.
The headless MWE span $w_i, \ldots, w_j$ has the corresponding tags $t_i = \text{B}$ and
$\forall k \in \{i+1, \ldots, j\}$, $t_k = \text{I}$;
tokens outside any spans are assigned the tag O.
We call $G$ and $T$ \emph{consistent} if they indicate the same set of headless spans for $w$.

\section{Three Approaches}
\label{sec:method}

We first present the standard approaches of edge-factored parsing (\refsec{sec:parsing}) and tagging (\refsec{sec:tagging}) for extracting headless spans in dependency trees,
and then introduce a joint decoder (\refsec{sec:joint-decoder}) that finds the global maximum among consistent (tree structure, tag sequence) pairs.

\subsection{Preliminaries}

Given a length-$n$ sentence $w$---which
we henceforth denote with the variable $x$ for consistency with machine-learning conventions---we first extract contextualized representations from the input
to associate each word with a vector
 $\bivec{x}_0$ (for the dummy word ``root''), $\bivec{x}_1$, \ldots, $\bivec{x}_n$.
We consider two common choices of feature extractors:
(1) bi-directional long short-term memory networks \citep[bi-LSTMs;][]{graves-schmidhuber05}
which have been widely adopted in dependency parsing \citep{kiperwasser-goldberg16b,dozat-manning17} and sequence tagging \citep{ma-hovy16a};
and
(2)  the Transformer-based \citep{vaswani+17} BERT feature extractor \citep{devlin+19}, pre-trained on large corpora
and  known to provide superior accuracies on both tasks \citep{kitaev+19,kondratyuk-straka19}.
For BERT models, we fine-tune the representations from the final layer
 for our parsing and tagging tasks.
When the BERT tokenizer renders multiple tokens from a single pre-tokenized word,
we follow \citet{kitaev+19} and use the BERT features from the last token as its representation.

\subsection{(Edge-Factored) Parsing}
\label{sec:parsing}

Since we consider headless structures that are embedded inside parse trees,
 it is natural to
identify them through a 
rule-based post-processing step after full parsing.
Our parsing component replicates that of the state-of-the-art \citet{che+18} parser,
which has the same parsing model as \citet{dozat-manning17}.
We treat unlabelled parsing as a head selection problem \citep{zhang+17} with deep biaffine attention scoring:
\begin{align*}
    \bivec{h}^{\text{attach}}_i &= \text{MLP}^{\text{attach-head}}(\bivec{x}_i)\\
    \bivec{m}^{\text{attach}}_j &= \text{MLP}^{\text{attach-mod}}(\bivec{x}_j)\\
    \textbf{s}_{i,j} &= [\bivec{h}^{\text{attach}}_i;1]^\top U^{\text{attach}} [\bivec{m}^{\text{attach}}_j;1]\\
    P(h_j=i\, |\, x) &= \text{softmax}_i(\bivec{s}_{:,j}),
\end{align*}
where $\text{MLP}^{\text{attach-head}}$ and $\text{MLP}^{\text{attach-mod}}$
are multi-layer perceptrons (MLPs)
that project contextualized representations into a $d$-dimensional space;
$[\cdot;1]$
indicates appending an extra entry of 1 to the vector;
$U^{\text{att}}\in \mathbb{R}^{(d+1)\times(d+1)}$ generates a score $s_{i,j}$ for $w_j$
attaching
to $w_i$
(which we can then refer to as the head of $w_j$, $h_j$);
a softmax function defines a probability distribution over all syntactic head candidates in the argument vector
(we use the range operator ``:'' to evoke a vector);
and, recall, we represent potential heads as integers, so that we may write
$h_j = i \in \{0, \ldots, n\}$.

The model for arc labeling employs an analogous deep biaffine scoring function:
\begin{align*}
    \bivec{h}^{\text{rel}}_i &= \text{MLP}^{\text{rel-head}}(\bivec{x}_i)\\
    \bivec{m}^{\text{rel}}_j &= \text{MLP}^{\text{rel-mod}}(\bivec{x}_j)\\
    \mathbf{v}_{i,j,r} &= [\bivec{h}^{\text{rel}}_i;1]^\top U^{\text{rel}}_r [\bivec{m}^{\text{rel}}_j;1]\\
    \hspace{-10pt}
    P(r_j=r\, |\, x,h_j=i) &= \text{softmax}_r(\bivec{v}_{i,j,:}),
\end{align*}
where $r_j$ is the arc label between $w_{h_j}$ and $w_j$.

The objective for training the parser is to minimize the  cumulative negative log-likelihood
\begin{align*}
L^{\text{parse}}=&\sum_{(i^*,j^*,r^*)\in E}[-\log P(h_{j^*}=i^*\, |\, x)\\&-\log P(r_i=r^*\, |\, x, h_{j^*}=i^*)].
\end{align*}
After the model predicts a full parse,
we extract headless structures as
the tokens ``covered'' by the longest-spanning $f$-arcs ($f=$~\deprel{flat} in UD).

\subsection{Tagging}
\label{sec:tagging}

For extracting spans in texts, if one chooses to ignore the existence of parse trees,
BIO tagging is a natural choice.
We treat the decision for the label of each token as an individual multi-class classification problem.
We let
\begin{equation*}
    P(t_i=t\, |\, x) = \text{softmax}_t(\text{MLP}^{\text{tag}}(\bivec{x}_i)),
\end{equation*}
where $\text{MLP}^{\text{tag}}$ has 3 output units corresponding to the scores for tags B, I and O respectively.\footnote{
    Sequence tagging is traditionally handled by conditional random fields \citep[][CRFs]{lafferty+01}.
    However, in recent experiments using contextualized representations on tagging \cite{clark+18,devlin+19},
    CRF-style loss functions provide little, if any,
    performance gains compared with simple 
    multi-class classification solutions, %
    at slower training speeds, to boot.
    Our preliminary experiments with both bi-LSTM and BERT-based encoders corroborate these findings,
    and thus we report results trained without CRFs.
}

We train the tagger to minimize
\begin{equation*}
    L^{\text{tag}}=\sum_{i}-\log P(t_i=t_i^*\, |\, x),
\end{equation*}
where $t^*$ corresponds to the gold BIO sequence.
During inference, we predict the BIO tags independently at each token position
and interpret the tag sequence as a set of MWE spans.
As a post-processing step, we discard all single-token spans,
since the task is to predict multi-word spans.

\subsection{A Joint Decoder}
\label{sec:joint-decoder}

\begin{figure*}[!ht]
    \begin{subfigure}{1.0\textwidth}
        \centering
        \small
        \begin{tabular}{rlrl}
            \multicolumn{2}{l}{\textbf{Axioms:}}
            &
            \\
            \multicolumn{4}{c}{
            \textsc{R-init}:\quad
            $\inferrule{ }{
                \tikz[baseline=-10pt]{\righttriangle[0.4][0.4]{i}{i}}:\colorbox{cyan!25}{$\log P(t_i=\text{O})$}
            }$
            \quad\quad
            \textsc{L-init}:\quad
            $\inferrule{ }{
                \tikz[baseline=-10pt]{\lefttriangle[0.4][0.4]{i}{i}}:0
            }$
            }
            \vspace{10pt}
            \\
            \multicolumn{4}{c}{
            \quad\quad
            \colorbox{cyan!25}{
            \textsc{R-mwe}:\quad
            $\inferrule{~}{
                \tikz[baseline=-10pt]{\righttriangle[0.4][0.4]{i}{j}}:\delta(i,j)
            }$,
            }
            }
            \\
            \multicolumn{4}{r}{
                \colorbox{cyan!25}{
                where $\delta(i,j)=\log P(t_i=\text{B}) + \sum_{k=i+1}^j{\left(\log P(t_k=\text{I}) + \log P(h_k=i)\right)}$
                }
            }
            \vspace{10pt}
            \\
            \multicolumn{2}{l}{\textbf{Deduction Rules:}}
            &
            \vspace{5pt}
            \\
            \textsc{R-comb}:
            &
            $\inferrule{
                \tikz[baseline=-15pt]{\trapezoid[0.6][0.6][0.3]{i}{k}}:s_1\quad
                \tikz[baseline=-10pt]{\righttriangle[0.5][0.4]{k}{j}}:s_2
            } {
                \tikz[baseline=-10pt]{\righttriangle[0.8][0.6]{i}{j}}:s_1+s_2
            }$
            &
            \textsc{R-link}:
            &
            $\inferrule{
                \tikz[baseline=-9pt]{\righttriangle[0.5][0.5]{i}{k}}:s_1\quad
                \tikz[baseline=-9pt]{\lefttriangle[0.5][0.5]{k+1}{j}}:s_2
            } {
                \tikz[baseline=-10pt]{\trapezoid[0.6][0.6][0.3]{i}{j}}:s_1+s_2+\log P(h_j=i)
            }$
            \\
            \textsc{L-comb}:
            &
            $\inferrule{
                \tikz[baseline=-9pt]{\lefttriangle[0.5][0.4]{j}{k}}:s_1\quad
                \tikz[baseline=-9pt]{\trapezoid[0.6][0.4][-0.3]{k}{i}}:s_2
            } {
                \tikz[baseline=-10pt]{\lefttriangle[0.8][0.6]{j}{i}}:s_1+s_2
            }$
            &
            \textsc{L-link}:
            &
            $\inferrule{
                \tikz[baseline=-9pt]{\righttriangle[0.5][0.5]{j}{k-1}}:s_1\quad
                \tikz[baseline=-9pt]{\lefttriangle[0.5][0.5]{k}{i}}:s_2\quad
            } {
                \tikz[baseline=-6pt]{\trapezoid[0.6][0.3][-0.3]{j}{i}}:s_1+s_2+\log P(h_j=i)
            }$
            \\
        \end{tabular}
    \end{subfigure}
    \caption{
     \posscite{eisner96} algorithm adapted to parsing headless structures
     (unlabeled case), 
     our modifications 
     highlighted in blue.
     All deduction items are annotated with their scores.
     \textsc{R-mwe} combines BIO tagging scores and head selection parsing scores.
     We need no \textsc{L-mwe} 
     because of the rightward headless-structure-arc convention.
    }
    \label{fig:algo}
\end{figure*}

A parser and a tagger take two different views of the same underlying data.
It is thus reasonable to hypothesize that a joint decoding process
that combines the scores from the two models might yield more accurate predictions.
In this section, we propose such a joint decoder to find the
parser+tagger-consistent structure with the highest product of probabilities.
Formally,
if $\mathcal{Y}$ is the output space for all consistent parse tree structures and BIO tag sequences,
for $y \in \mathcal{Y}$
with components  consisting of tags $t_i$,  head assignments $h_i$,  and relation labels $r_i$,
our decoder aims to find $\hat{y}$ satisfying
\begin{equation*}
    \hat{y} = \arg\max_{y\in\mathcal{Y}}{P(y\, |\, x)},
\end{equation*}
where
\begin{equation*}
    P(y\, |\, x) = \prod_i{P(t_i\, |\, x)P(h_i\, |\, x)P(r_i\, |\, x,h_i)}.
\end{equation*}

\reffig{fig:algo} illustrates our joint decoder in the unlabeled case.\footnote{
    In the labeled case, the parser further adds the arc-labeling scores to the \textsc{R-mwe} and \textsc{link} rules.
}
It builds on \posscite{eisner96} decoder for projective dependency parsing.
In addition to having single-word spans as axioms in the deduction system,
we further allow multi-word spans to enter the decoding procedures through the axiom \textsc{R-mwe}.
Any initial single-word spans receive an O-tag score for that word,
while the newly introduced MWE spans receive B-tag, I-tag,
attachment and relation scores that correspond to the two consistent views of the same structure.
The time complexity for this decoding algorithm remains the same $O(n^3)$ as the original \citeauthor{eisner96} algorithm.

During training, we let the parser and the tagger share the same contextualized representation $\bivec{x}$
and optimize a linearly interpolated joint objective
\begin{equation*}
    L^{\text{joint}}=\lambda L^{\text{parse}} + (1-\lambda) L^{\text{tag}},
\end{equation*}
where $\lambda$ is a hyper-parameter adjusting the relative weight of each module.\footnote{
    The joint decoder combines tagging and parsing scores regardless of whether the two modules are jointly trained.
    However, since feature extraction is the most time-consuming step in our neural models,
    especially with BERT-based feature extractors,
    it is most practical to save memory and time by sharing common feature representations across modules.
}
This is an instance of multi-task learning \citep[MTL;][]{caruana93,caruana97}.
MTL
 has proven
to be a successful technique \citep{collobert-weston08}
 on its own;
thus, in our experiments, we compare the joint decoder and using the MTL strategy alone.

\section{Experiments}
\label{sec:exp}

\begin{table*}[ht]
  \centering
    \begin{tabular}{llrrrrcr}
    \toprule
        \multicolumn{2}{l}{\multirow{2}{*}{Treebank}} & \multirow{2}{*}{\# tokens} & \multicolumn{1}{c}{\# headless} & 
        \multirow{2}{*}{\% \hspace*{.25cm}}  & \multicolumn{1}{c}{\# headless} & \multicolumn{1}{c}{Average}     & \multicolumn{1}{c}{Compliance} \\
                          &    &                               & \multicolumn{1}{c}{arcs}        &   & \multicolumn{1}{c}{spans}       & \multicolumn{1}{c}{span length} & \multicolumn{1}{c}{ratio} \\
    \midrule
        \multicolumn{2}{l}{English} & $731{,}677$ & $32{,}065$ & $4.38\%$ & $16{,}997$ & $2.89$ & $100.00\%$ \\
        \parbox[t]{0mm}{\multirow{6}{*}{\rotatebox[origin=c]{90}{\rule{20pt}{0.4pt}~UD 2.2~\rule{20pt}{0.4pt}}}}
        &\treebank{de\_gsd}        & $263{,}804$ & $6{,}786$ & $2.57\%$ & $5{,}663$ & $2.59$ & $93.00\%$  \\
        &\treebank{it\_postwita}   & $99{,}441$  & $2{,}733$ & $2.75\%$ & $2{,}277$ & $2.26$ & $94.89\%$  \\
        &\treebank{nl\_alpino}     & $186{,}046$ & $4{,}734$ & $2.54\%$ & $3{,}269$ & $2.45$ & $100.00\%$ \\
        &\treebank{nl\_lassysmall} & $75{,}134$  & $4{,}408$ & $5.87\%$ & $3{,}018$ & $2.46$ & $99.82\%$  \\
        &\treebank{no\_nynorsk}    & $245{,}330$ & $5{,}578$ & $2.27\%$ & $3{,}670$ & $2.54$ & $99.78\%$  \\
        &\treebank{pt\_bosque}     & $206{,}739$ & $5{,}375$ & $2.60\%$ & $4{,}310$ & $2.25$ & $97.38\%$  \\
    \bottomrule
    \end{tabular}
  \caption{
    Dataset statistics.
    Language codes: \treebank{de}=German; \treebank{it}=Italian; \treebank{nl}=Dutch; \treebank{no}=Norwegian; \treebank{pt}=Portuguese.
  }
  \label{tab:data}%
\end{table*}

\begin{table*}[ht]
  \centering
    \begin{tabular}{ll|r|cc|cc|c}
    \toprule
        \multicolumn{2}{l|}{\underline{\emph{w/ bi-LSTM}}} & Compl. & \multicolumn{2}{c|}{}                              & \multicolumn{2}{c|}{MTL}                        & Joint\\
        \multicolumn{2}{l|}{Treebank} & Ratio $\downarrow$ & \multicolumn{1}{c}{Parsing} & \multicolumn{1}{c|}{Tagging} & \multicolumn{1}{c}{Parsing} & \multicolumn{1}{c|}{Tagging} & Decoding \\
    \midrule
        \multicolumn{2}{l|}{English}
                                   & $100.00$ & \resnumber{91.24}{0.60} & \resnumber{91.81}{0.45} & \resnumber{93.00}{0.83} & \close{93.24}{0.76}     & \best{93.49}{0.43}      \\
        \parbox[t]{0mm}{\multirow{6}{*}{\rotatebox[origin=c]{90}{\rule{20pt}{0.4pt}~UD 2.2~\rule{20pt}{0.4pt}}}}
        &\treebank{nl\_alpino}     & $100.00$ & \resnumber{72.66}{1.73} & \resnumber{74.94}{1.00} & \resnumber{77.29}{0.80} & \resnumber{75.58}{1.18} & \best{79.65}{1.05}      \\
        &\treebank{nl\_lassysmall}  & $99.82$ & \resnumber{76.44}{1.56} & \close{77.98}{1.56}     & \close{78.13}{0.98}     & \resnumber{77.58}{1.17} & \best{78.92}{1.00}      \\
        &\treebank{no\_nynorsk}     & $99.78$ & \resnumber{85.34}{0.81} & \resnumber{87.67}{0.90} & \resnumber{86.72}{0.76} & \resnumber{87.44}{0.76} & \best{88.40}{0.39}      \\
        &\treebank{pt\_bosque}      & $97.38$ & \resnumber{89.55}{1.10} & \resnumber{90.97}{0.46} & \close{91.30}{0.75}     & \best{92.07}{1.04}      & \resnumber{90.63}{1.56} \\
        &\treebank{it\_postwita}    & $94.89$ & \resnumber{75.35}{1.05} & \resnumber{76.37}{1.72} & \best{78.46}{1.08}      & \close{77.87}{0.57}     & \close{78.38}{1.04}     \\
        &\treebank{de\_gsd}         & $93.00$ & \resnumber{63.32}{1.36} & \resnumber{64.10}{1.31} & \close{64.81}{2.05}     & \close{65.07}{1.35}     & \best{65.86}{1.34}      \\
    \midrule
        \multicolumn{3}{c|}{Average}              & $79.13$                 & $80.55$                 & $81.39$                 & $81.26$                 & $82.19$                \\
    \bottomrule
    \end{tabular}

    \vspace{6pt}

    \begin{tabular}{ll|r|cc|cc|c}
    \toprule
        \multicolumn{2}{l|}{\underline{\emph{w/ BERT}}} & Compl. & \multicolumn{2}{c|}{}                              & \multicolumn{2}{c|}{MTL}                        & Joint\\
        \multicolumn{2}{l|}{Treebank}     & Ratio $\downarrow$ & \multicolumn{1}{c}{Parsing} & \multicolumn{1}{c|}{Tagging} & \multicolumn{1}{c}{Parsing} & \multicolumn{1}{c|}{Tagging} & Decoding \\
    \midrule
        \multicolumn{2}{l|}{English}
                                   & $100.00$ & \resnumber{94.98}{0.26} & \resnumber{95.45}{0.23} & \resnumber{95.01}{0.20} & \best{95.86}{0.19}      & \resnumber{95.51}{0.58} \\
        \parbox[t]{0mm}{\multirow{6}{*}{\rotatebox[origin=c]{90}{\rule{20pt}{0.4pt}~UD 2.2~\rule{20pt}{0.4pt}}}}
        &\treebank{nl\_alpino}     & $100.00$ & \resnumber{83.87}{1.61} & \resnumber{83.32}{1.01} & \resnumber{84.65}{1.48} & \close{85.90}{1.51}     & \best{86.61}{1.52}      \\
        &\treebank{nl\_lassysmall}  & $99.82$ & \resnumber{87.16}{1.20} & \resnumber{87.52}{0.59} & \close{88.10}{0.80}     & \resnumber{87.68}{0.78} & \best{88.35}{0.49}      \\
        &\treebank{no\_nynorsk}     & $99.78$ & \resnumber{92.16}{0.93} & \best{93.48}{0.48}      & \resnumber{92.45}{0.34} & \close{93.11}{0.21}     & \close{93.08}{0.62}     \\
        &\treebank{pt\_bosque}      & $97.38$ & \resnumber{92.98}{0.82} & \resnumber{93.47}{0.55} & \resnumber{93.42}{0.65} & \close{93.85}{0.57}     & \best{94.01}{0.19}      \\
        &\treebank{it\_postwita}    & $94.89$ & \resnumber{80.80}{1.51} & \resnumber{80.80}{1.52} & \close{80.90}{1.78}     & \best{81.33}{0.43}      & \resnumber{80.83}{1.20} \\
        &\treebank{de\_gsd}         & $93.00$ & \resnumber{68.21}{1.43} & \close{70.28}{0.70}     & \close{70.04}{1.14}     & \best{71.05}{1.12}      & \close{70.72}{0.90}     \\
    \midrule
        \multicolumn{3}{c|}{Average}              & $85.74$                 & $86.33$                 & $86.37$                 & $86.97$                 & $87.02$                \\
    \bottomrule
    \end{tabular}
  \caption{
    Flat-structure identification test-set F1 scores (\%)  with bi-LSTM (top) and BERT (bottom).
    The cell with the best result for each treebank has blue shading; results within one  standard deviation of the best are bolded.
  }
  \label{tab:main-results}%
\end{table*}

\paragraph{Data}
\label{sec:data}
We perform experiments on the MWE-Aware English Dependency Corpus \cite{kato+17}
and treebanks selected from Universal Dependencies 2.2 \cite[UD;][]{ud22} for
having frequent occurrences of headless MWE structures.
The MWE-Aware English Dependency Corpus provides
automatically unified named-entity annotations based on OntoNotes 5.0 \cite{weischedel:2013:ontonotes5}
and Stanford-style dependency trees \cite{demarneffe-manning08}.
We extract MWE spans according to \deprel{mwe\_NNP} dependency relations.
We choose the UD treebanks based on
two basic properties that hold for flat structures conforming
to the UD annotation guidelines:
(1) all words that are attached via \deprel{flat} relations must be leaf nodes
and (2) all words within a \deprel{flat} span should be attached to a common ``head'' word, and each arc label should be either \deprel{flat} or \deprel{punct}.\footnote{
\deprel{punct} inside a headless span is often used for hyphens and other internal
punctuation in named entities. See the English sentence in \reffig{fig:ud-ex} for an example.
}
For each treebank, we compute its \emph{compliance ratio}, defined as the
percentage of its trees containing \deprel{flat} arc labels that satisfy
both properties above; and we filter out those with compliance ratios below 90\%.\footnote{
The two properties defined in the UD guidelines for headless structures provide us with
a common basis for uniform treatment across languages and treebanks.
Unfortunately, the two properties can be violated quite often, due to issues in
annotation and automatic treebank conversion into UD
style.
In $6$ out of the top $10$ treebanks containing the most \deprel{flat} relations,
(at least one of) these properties are violated in more than $35\%$ of the sentences with \deprel{flat} relations
and have to be excluded from our experiments.
We
hope that
ongoing community effort in data curation will facilitate evaluation on more diverse languages.
}
We rank the remaining treebanks by their ratios of \deprel{flat} relations among all dependency arcs,
and pick those with ratios higher than 2\%.
Six treebanks representing 5 languages,
German \cite{mcdonald+13}, Italian \cite{sanguinetti+18}, Dutch \cite{bouma-vannoord17}, Norwegian \cite{solberg+14} and Portuguese \cite{rademaker+17},
are selected for our experiments.\footnote{
It is a coincidence that all the selected languages are Indo-European (IE).
Although there are some non-IE treebanks with high \deprel{flat} ratio, such as Korean
(see \reftab{tab:ud-basic-stats}),
the annotated structures frequently break one or both of the basic properties.
See \reffig{fig:ud-ex} for violation examples.
}
Data statistics are given in \reftab{tab:data}.
To construct gold-standard BIO labels,
we extract MWE spans according to the longest-spanning arcs that correspond to headless structures.

\paragraph{Implementation Details}

We use $3$-layer bi-LSTMs
where each layer has $400$ dimensions in both directions
and the inputs are concatenations of $100$-dimensional randomly-initialized word embeddings
with
the
final hidden vectors of $256$-dimensional single-layer character-based bi-LSTMs;
for BERT,
we use pre-trained cased multi-lingual BERT models\footnote{
    \url{https://github.com/huggingface/transformers}
}
and fine-tune the weights.
We
adopt
the parameter settings
of
\citet{dozat-manning17} and
use 500 and 100 dimensions for $U^{\text{att} }$ and $U^{\text{rel}}_r$, respectively.
The MLP in the taggers have $500$ hidden dimensions.
We use a dropout \cite{srivastava+14} rate of $0.33$, a
single hidden layer,
and a ReLU activation function \citep{nair-hinton10} for all MLPs.
The models are trained with
the
 Adam optimizer \cite{kingma-ba15} using
a
 batch size of $16$ sentences.
The learning rates are set to $1e^{-3}$ for bi-LSTMs and $1e^{-5}$ for BERT initially
and then multiplied by a factor of $0.1$ if the performance on the development set stops improving within $3200$ training iterations.
For the parsing models, we use the projective
\citet{eisner96} decoder algorithm.
For the joint training and joint decoding models,
we tune $\lambda\in\{0.02,0.05,0.1,0.3,0.5,0.9\}$ for each treebank independently
and fix the settings based on the best dev-set scores.
We run each model with $5$ different random seeds and report the mean and standard deviation for each setting.

\paragraph{Results}
We
report
F1 scores based on multi-word headless-structure extraction.
\reftab{tab:main-results} compares different strategies for identifying headless MWEs in parse trees.
Tagging is consistently better than parsing except for two treebanks with BERT feature extractor.
Tagging beats parsing in all but two combinations of treebank and feature extractor.
As hypothesized, our joint decoder improves over both strategies by $0.69\%$ ($1.64\%$) absolute
through combined decisions from parsing and tagging with(out) BERT.
We also compare the joint decoding setting with MTL training strategy alone.
While joint decoding yields superior F1 scores, MTL is responsible for a large portion of the gains:
it accounts for over half of the average gains with bi-LSTMs,
and when we use pre-trained BERT feature extractors,
the accuracies of jointly-trained taggers
are essentially as good as
joint decoding models.

Interestingly, the choice of feature extractors also has an effect on the performance gap between parsers and taggers.
With bi-LSTMs, tagging is $1.42\%$ absolute F1 higher than parsing, and the gap is mitigated through MTL.
While pre-trained BERT reduces the performance difference dramatically down to $0.59\%$ absolute,
MTL no longer helps parsers overcome this gap.
Additionally, we observe that MTL helps
both parsing and tagging models,
demonstrating that the two views of the same underlying structures are complementary to each other and that learning both can be beneficial to model training.
By resolving such representational discrepancies, joint decoding exhibits further accuracy improvement.

In terms of dependency parsing accuracies,
we confirm that our parsing-only models achieve state-of-the-art performance on the UD treebanks,
but there are no significant differences in parsing results among parsing-only, MTL and jointly-decoded models.
See Appendix for detailed results.

\section{Related Work}
\label{sec:related}

Syntactic analysis in conjunction with MWE identification is an important line of research \cite{wehrli00}.
The span-based representations 
that form the basis of
phrase-structure trees 
(as opposed to dependency trees)
are arguably directly compatible with headless spans.
This motivates approaches using joint constituency-tree representations based on context-free grammars \cite{arun-keller05,constant+13}
and tree substitution grammars \cite{green+11,green+13}.
\citet{finkel-manning09} add new phrasal nodes to denote named entities,
enabling statistical parsers trained on this modified representation to produce both parse trees and named entity spans simultaneously.
\citet{leroux+14} use dual decomposition to develop a joint system that combines phrase-structure parsers and taggers for compound recognition.
These approaches do not directly transfer to dependency-based representations
since dependency trees do not explicitly represent phrases.

In the context of dependency parsing, \citet{eryigit+11} report that MWE annotations have a large impact on parsing.
They find that the dependency parsers are more accurate when MWE spans are not unified into single lexical items.
Similar to the phrase-structure case, \citet{candito-constant14} consider MWE identification as a side product of dependency parsing into joint representations.
This parse-then-extract strategy is widely adopted \cite{vincze+13,nasr+15,simko+17}.
\citet{waszczuk+19} introduce additional parameterized scoring functions for the arc labelers
and use global decoding to produce consistent structures during arc-labeling steps once unlabeled dependency parse trees are predicted.
Our work additionally proposes a joint decoder that combines the scores from both parsers and taggers.
Alternative approaches to graph-based joint parsing and MWE identification
include transition-based \cite{constant-nivre16}
and easy-first \cite{constant+16} dependency parsing.
These approaches typically rely on greedy decoding,
whereas our joint decoder finds the 
globally optimal solution through dynamic programming.

Our work only focuses on a subset of MWEs that do not have internal structures.
There 
is substantial research interest
in the broad area of MWEs \cite{sag02,constant+17}
including recent releases of datasets \citep{schneider-smith15}, editions of shared tasks \citep{savary+17,ramisch+18} and workshops \citep{savary+18,savary+19}.
We leave it to future work to extend the comparison and combination 
of
taggers and dependency parsers to other MWE constructions.

\section{Conclusion and Further Directions}
\label{sec:conclusion}

Our paper provides an empirical comparison of different strategies for extracting headless MWEs from dependency parse trees:
parsing, tagging, and joint modeling.
Experiments on the MWE-Aware English Dependency Corpus and UD 2.2 across five languages show that
tagging, a widely-used methodology for extracting spans from texts, is more accurate than parsing for this task.
When using bi-LSTM (but not BERT) representations, 
our proposed joint decoder reaches higher F1 scores than either 
of the two other strategies, 
by combining scores of the two different and complementary representations of the same structures.
We also show that most of the gains stem from a multi-task learning strategy
that shares common neural representations between the parsers and the taggers.

An interesting 
additional use-case
for
our
joint decoder is when a downstream task,
e.g., relation extraction, requires output structures from both a parser and a tagger.
Our joint decoder can find the highest-scoring consistent structures among all candidates,
and thus 
has the potential 
to provide
simpler model designs in downstream applications.

Our study has been limited to a few treebanks in UD partially due to large variations and inconsistencies across different treebanks.
Future community efforts on a unified representation of flat structures for all languages would facilitate further research on
linguistically-motivated treatments of headless structures in ``headful'' dependency treebanks.

Another limitation of our current work is that our joint decoder only produces projective dependency parse trees.
To handle non-projectivity, one possible solution is pseudo-projective parsing \cite{nivre-nilsson05}.
We leave it to future work to design a non-projective decoder for joint parsing and headless structure extraction.

\section*{Acknowledgments}

We thank the three anonymous reviewers for their comments,
and
Igor Malioutov, Ana Smith and the Cornell NLP group
for discussion and comments.
TS was supported by a Bloomberg Data Science Ph.D. Fellowship.

\bibliography{ref}
\bibliographystyle{acl_natbib}

\clearpage
\begin{appendices}
\setcounter{table}{0}
\renewcommand{\thetable}{A\arabic{table}}

\section{Evaluation of the Strengths of Our Parsing Models}

To confirm that we work with reasonable parsing models, we compare our parsers with those in the CoNLL 2018 shared task \cite{zeman+18}.
The shared task featured an end-to-end parsing task, requiring all levels of text processing including tokenization, POS tagging, 
morphological
analysis, etc.
We focus on the parsing task only, and predict syntactic trees based on sentences tokenized by the \citet{qi+18} submission.\footnote{
We thank the shared task participants and the organizers for making system predictions available at \url{https://lindat.mff.cuni.cz/repository/xmlui/handle/11234/1-2885}.
}
\reftab{tab:conll} shows that our parsing models are highly competitive with the current state-of-the-art.
Indeed, on four out of the six treebanks we selected for their density of flat structures,
our baseline models actually achieve higher labeled attachment scores (LAS) than the 
the top scorer did in the official shared task.

\begin{table}[ht]
  \centering
    \begin{tabular}{@{\hspace{3pt}}l@{\hspace{4pt}}c@{\hspace{8pt}}c@{\hspace{3pt}}}
\toprule
        \multirow{2}{*}{Treebank}    & Our & \multirow{1}{*}{CoNLL 2018} \\
             &  Parsers & Best \\
\midrule
\treebank{de\_gsd}        & $\mathbf{80.65}$ & $80.36$           \\
\treebank{it\_ostwita}    & $79.33$          & $\textbf{79.39}$  \\
\treebank{nl\_alpino}     & $\mathbf{89.78}$ & $89.56$           \\
\treebank{nl\_lassysmall} & $\mathbf{87.96}$ & $86.84$           \\
\treebank{no\_nynorsk}    & $90.44$          & $\textbf{90.99}$  \\
\treebank{pt\_bosque}     & $\mathbf{89.25}$ & $87.81$           \\
\bottomrule
\end{tabular}
  \caption{
      Comparison of our (non-MTL) parsing models with the best-performing systems \citep{che+18,qi+18} from the CoNLL 2018 shared task,
      measured by labeled attachment scores (LAS, \%).
  }
  \label{tab:conll}%
\end{table}

\newpage

\section{Do MTL and Joint Decoding Help Parsing Performance?}
{

\begin{table*}[t]
  \centering
    \begin{tabular}{ll|r|c|c|c}
    \toprule
        \multicolumn{2}{l|}{\underline{\emph{w/ bi-LSTM}}} & Compl. & \multicolumn{1}{c|}{}                              & \multicolumn{1}{c|}{MTL}                        & Joint\\
        \multicolumn{2}{l|}{Treebank} & Ratio $\downarrow$ & \multicolumn{1}{c|}{Parsing} & \multicolumn{1}{c}{Parsing}  & Decoding \\
    \midrule
        \multicolumn{2}{l|}{English}
                                   & $100.00$ & \close{89.30}{0.41}     & \close{89.39}{0.67}     & \best{89.77}{0.52}      \\
        \parbox[t]{0mm}{\multirow{6}{*}{\rotatebox[origin=c]{90}{\rule{20pt}{0.4pt}~UD 2.2~\rule{20pt}{0.4pt}}}}
        &\treebank{nl\_alpino}     & $100.00$ & \resnumber{81.97}{1.27} & \close{82.57}{0.99}     & \best{82.79}{0.77}      \\
        &\treebank{nl\_lassysmall}  & $99.82$ & \resnumber{82.06}{1.30} & \best{82.90}{0.64}      & \resnumber{81.55}{1.26} \\
        &\treebank{no\_nynorsk}     & $99.78$ & \close{86.54}{0.50}     & \close{86.35}{0.37}     & \best{86.65}{0.64}      \\
        &\treebank{pt\_bosque}      & $97.38$ & \resnumber{84.29}{2.15} & \resnumber{84.48}{1.61} & \best{85.28}{0.25}      \\
        &\treebank{it\_postwita}    & $94.89$ & \best{77.39}{0.69}      & \close{76.75}{1.29}     & \resnumber{76.59}{1.46} \\
        &\treebank{de\_gsd}         & $93.00$ & \best{76.66}{0.64}      & \close{76.35}{0.83}     & \resnumber{75.22}{1.98} \\
    \midrule
        \multicolumn{3}{c|}{Average}              & $82.60$             & $82.69$                 & $82.55$                \\
    \bottomrule
    \end{tabular}

    \vspace{6pt}

    \begin{tabular}{ll|r|c|c|c}
    \toprule
        \multicolumn{2}{l|}{\underline{\emph{w/ BERT}}} & Compl. & \multicolumn{1}{c|}{}                              & \multicolumn{1}{c|}{MTL}                        & Joint\\
        \multicolumn{2}{l|}{Treebank} & Ratio $\downarrow$ & \multicolumn{1}{c|}{Parsing} & \multicolumn{1}{c}{Parsing}  & Decoding \\
    \midrule
        \multicolumn{2}{l|}{English}
                                   & $100.00$ & \best{93.73}{0.24}      & \close{93.52}{0.17}     & \resnumber{93.38}{0.39} \\
        \parbox[t]{0mm}{\multirow{6}{*}{\rotatebox[origin=c]{90}{\rule{20pt}{0.4pt}~UD 2.2~\rule{20pt}{0.4pt}}}}
        &\treebank{nl\_alpino}     & $100.00$ & \close{89.82}{0.55}     & \best{89.95}{0.41}      & \close{89.86}{0.59}     \\
        &\treebank{nl\_lassysmall}  & $99.82$ & \best{89.78}{0.46}      & \close{89.76}{0.17}     & \close{89.67}{0.16}     \\
        &\treebank{no\_nynorsk}     & $99.78$ & \close{90.77}{0.20}     & \best{90.98}{0.38}      & \close{90.85}{0.32}     \\
        &\treebank{pt\_bosque}      & $97.38$ & \close{89.78}{0.32}     & \close{89.51}{0.39}     & \best{89.79}{0.39}      \\
        &\treebank{it\_postwita}    & $94.89$ & \close{81.61}{0.32}     & \best{81.70}{0.14}      & \resnumber{81.53}{0.63} \\
        &\treebank{de\_gsd}         & $93.00$ & \resnumber{81.51}{0.23} & \best{81.74}{0.23}      & \close{81.52}{0.17}     \\
    \midrule
        \multicolumn{3}{c|}{Average}              & $88.14$             & $88.17$                 & $88.09$                \\
    \bottomrule
    \end{tabular}
  \caption{
      Dependency-parsing labeled attachment scores (LAS, \%) on the test sets with bi-LSTM (top) and BERT (bottom) feature extractors.
      The cell containing the best result for each treebank has blue shading; results within one standard deviation of the best are in boldface.
  }
  \label{tab:parse-results}%
\end{table*}

}

In \reftab{tab:parse-results} (next page), we investigate whether 
MTL
and combining scores from both representations of flat-structure MWEs
can
improve parsing performance.
We observe very little difference among the various strategies.
This fact can be explained by the relatively low ratios of flat relations and the already-high base performance:
the room for 
improvement on the standard LAS metrics is quite small.

\end{appendices}

\end{document}